\def\year{2018}\relax
\documentclass[letterpaper]{article} 
\usepackage{aaai18}  
\usepackage{times}  
\usepackage{helvet}  
\usepackage{courier}  
\usepackage{url}  
\usepackage{graphicx}  
\usepackage{subfigure}
\usepackage{amsmath}
\usepackage{amsfonts}
\usepackage{amssymb}
\usepackage{mathrsfs}

\begin{document}
%
\title{f-VAEs: Improve VAEs with Conditional Flows}
\author{Jianlin Su \ \qquad\qquad\qquad\qquad\qquad\qquad Guang Wu\\
School of Mathematics, Sun Yat-sen University
\ \qquad School of Hefei University of Technology
}
\maketitle
\begin{abstract}
In this paper, we integrate VAEs and flow-based generative models successfully and get f-VAEs. Compared with VAEs, f-VAEs generate more vivid images, solved the blurred-image problem of VAEs. Compared with flow-based models such as Glow, f-VAE is more lightweight and converges faster, achieving the same performance under smaller-size architecture.
\end{abstract}

\noindent Recently, deep generative models has been widely studied and developed. Outside of Generative Adversarial
Networks (GANs) \cite{Goodfellow2014Generative}, Variational Autoencoders (VAEs) \cite{Kingma2013Auto} and flow-based models \cite{Dinh2014NICE,Dinh2016Density} are two distinct kinds of competitive generative models. They have their own advantages and disadvantages, and we try to integrate them to a new model.

\section{Foundation}

Let $\tilde{p}(x)$ be the evidence distribution of the dataset. The basic idea of generative models is to fit the dataset by the following formulation distribution
\begin{equation}q(x)=\int q(z)q(x|z) dz\end{equation}
whose prior distribution $q(z)$ is standard Gaussian distribution usually. $q(x|z)$ discribes a generative procedure, which is conditional Gaussian distribution (in VAEs) or Delta $\delta$ distribution (in GANs and flow-based models). Ideally, we maximize log likelihood function $\mathbb{E}[\log q(x)]$ , or equally, minimize KL divergence $KL(\tilde{p}(x)\Vert q(x))$, to find out the best model.

However, the integral usually has no explicit result, so we need some trick solutions. One of them is VAEs, who introduces a posterior distribution $p(z|x)$, and change to minimize joint KL divergence $KL(\tilde{p}(x)p(z|x)\Vert q(z)q(x|z))$, which is an upper bound of $KL(\tilde{p}(x)\Vert q(x))$ and always easy-to-compute \cite{Su2018Variational}.

Another solution is flow-based models, who let $q(x|z)=\delta(x - G(z))$ and calculate the integral out by well designed $G(z)$ (stacks of flow). The major component is Coupling layer: split $x$ into two parts $x_1,x_2$ and then
\begin{equation}\label{eq:coupling}\begin{aligned}&y_1 = x_1\\
&y_2 = s(x_1)\otimes x_2 + t(x_1)
\end{aligned}\end{equation}
whose inverse is
\begin{equation}\begin{aligned}&x_1 = y_1\\
&x_2 = (y_2 - t(y_1)) /  s(x_1)
\end{aligned}\end{equation}
and Jacobi determinant is $\prod_i s_i(x_i)$. This transformation we call it \emph{Affine Coupling} (if $s(x_1)\equiv 1$, it's also called \emph{Additive Coupling}), denoted as $f$. Many Coupling layers can be combined to generate a complex invertible function, that is $G = f_1 \circ f_2 \circ \dots \circ f_n$, called (unconditional) \emph{flow}. 

Therefore flow-based models can maximize log likelihood function directly. Recently, Glow model \cite{Kingma2018Glow} demonstrates the very realistic generative effect of flow-based models, arousing many people's interest in flow-based models again. But flow-based models is usually very heavyweight, for example, the Glow model for 256x256 images needs to train for a week on 40 GPUs \footnote{the details can be found at \url{https://github.com/openai/glow/issues/14#issuecomment-406650950} and \url{https://github.com/openai/glow/issues/37#issuecomment-410019221}.}. So flow-based model is not friendly enough.

\section{Analysis}

As we know, the images generated by VAEs always will blur. Some thinks it is the consequence of MSE loss, some thinks it is the inherent problem of KL divergence. However, one we need to know is that the reconstructed images by ordinary AutoEncoders are always blurring too. So the blurring problem may be an inherent problem of AE, which need to reconstruct high-dimensional data by low-dimensional hidden variable.

How about if the size of hidden variable equals the input image size? Not enough. Because of the Gaussian assumption of $p(z|x)$, the fitting ability of VAEs is limited, while the Gaussian distributions is just a tiny subset of all possible posterior distributions.

So what is the problem of flow-based models? flow-based models design a invertible (and strongly nonlinear) transformation to encode input distribution to Gaussian. For this transformation, we have to guarantee not only the invertibility but also the computability of Jacobi determinant, which leads to $\eqref{eq:coupling}$. However, these Coupling layers can only generate very weak nonlinearity, so we need to stack a lot of Coupling layers to achieve strong nonlinearity, which leads flow-based models to be very heavy.

\section{f-VAEs}

Our new model is to introduce flow into VAEs, using flow-based model to construct more powerful posterior distribution in stead of Gaussian ditribution. We called it \emph{Flow-based Variational Autoencoders} (f-VAEs). It can generate clear images (compared with VAE) and has a lighter weight (compared with Glow).

\subsection{Derivation}

We start from the original loss of VAE:
\begin{equation}\label{eq:vae-loss} \begin{aligned}&KL(\tilde{p}(x)p(z|x)\Vert q(z)q(x|z))\\
=&\iint \tilde{p}(x)p(z|x)\log \frac{\tilde{p}(x)p(z|x)}{q(x|z)q(z)} dzdx\end{aligned}\end{equation}
different with standard VAE, we no longer assume that $p(z|x)$ is Gaussian, instead of a flow-based formulation
\begin{equation}\label{eq:cond-flow} p(z|x) = \int \delta(z - F_x(u))q(u)du\end{equation}
here $q(u)$ is standard Gaussian distribution. $F_x(u)$ is a bivariate function of $x,u$ and invertible for variable $u$. We can regard it as a flow-based model of $u$, but its parameters are the function of variable $x$. Here we call it \emph{conditional flow}. Replace $p(z|x)$ in $\eqref{eq:vae-loss}$ with it, we get
\begin{equation}\label{eq:f-vae-loss}\iint \tilde{p}(x)q(u)\log \frac{\tilde{p}(x) q(u)}{q(x| F_x(u))q(F_x(u))\left|\det \left[\frac{\partial F_x (u)}{\partial u}\right]\right|} dudx\end{equation}
that is the general loss of f-VAE. The details can be found at Appenix A.

\subsection{Two Cases}

Now we calculate two special simple cases to see how $\eqref{eq:f-vae-loss}$ works. Firstly, we let
\begin{equation}\label{eq:vae-fxu} F_x(u)=\sigma(x)\otimes u + \mu(x)\end{equation}
so we have
\begin{equation}-\int q(u)\log \left|\det \left[\frac{\partial F_x (u)}{\partial u}\right]\right| du=-\sum_i\log \sigma_i(x)\end{equation}
and
\begin{equation}\int q(u)\log \frac{q(u)}{q(F_x(u))}du=\frac{1}{2}\sum_{i=1}^d(\mu_i^2(x)+\sigma_i^2(x)-1)\end{equation}
The sum of them equals $KL(p(z|x)\Vert q(z))$. Plug it into $\eqref{eq:f-vae-loss}$, we get the loss of standard VAE, consisting reparameterization trick automatically.

Secondly, we choose
\begin{equation}\label{eq:flow-fxu} F_x(u)=F(\sigma u + x),\quad q(x|z)=\mathscr{N}(x;F^{-1}(z),\sigma)\end{equation}
while $\sigma$ is a small positive constant and $F$ is a flow-based encoder whose parameters are independent of $x$ (unconditional flow). Then we can see
\begin{equation}\begin{aligned}&-\log q(x|F_x(u))\\
=& -\log \mathscr{N}(x; F^{-1}(F(\sigma u + x)),\sigma)\\
=& -\log \mathscr{N}(x; \sigma u + x,\sigma)\\
=& \frac{d}{2}\log 2\pi \sigma^2 + \frac{1}{2}\Vert u\Vert^2
\end{aligned}\end{equation}
It means there is no training parameters in $\log q(x|F_x(u))$. Therefore, the total loss equals
\begin{equation}-\iint \tilde{p}(x)q(u)\log q(F(\sigma u + x))\left|\det \left[\frac{\partial F(\sigma u + x)}{\partial u}\right]\right| dudx\end{equation}
That is the loss of original flow-based models, whose inputs are added Gaussian noise with variance $\sigma^2$. Interestingly, the standard training strategy of flow-based models will actually add some noise into inputs and our result includes it naturally.

\subsection{Our Model}

From above we can see the original VAEs and flow-based models are included in $\eqref{eq:f-vae-loss}$ natrually. $F_x(u)$ discribes how the information of $x$ and $U$ mix up, and in theory we can use any complicated $F_x(u)$ to improve fitting ability of $p(z|x)$, for example,
\begin{equation}\begin{aligned}&f_1 = F_1\Big(\sigma_1(x)\otimes u + \mu_1(x)\Big)\\
&f_2 = F_2\Big(\sigma_2(x)\otimes f_1 + \mu_2(x)\Big)\\
&F_x(u) = \sigma_3(x)\otimes f_2 + \mu_3(x)\end{aligned}\end{equation}
whose $F_1, F_2$ are unconditional flow.

Up to now we do not have any constraint for the size of $z$ (or $u$). Actually it is a hyperparameter in the model. Thus $\eqref{eq:f-vae-loss}$ allow us to train a better dimension-reduced VAE or dimension-reduced flow-based models. Howerver, for image generating, we have seen that dimension-reduced AutoEncoders may lead to blurring problem. So in this paper, we let the size of $z$ equals the size of input image $x$.

Out of practicability and simplicity, we combine $\eqref{eq:flow-fxu}$ and $\eqref{eq:vae-fxu}$ to a general formulation
\begin{equation}\label{eq:f-vae-fxu} \begin{aligned}F_x(u)=&F(\sigma_1 u + E(x)),\\
q(x|z)=&\mathscr{N}(x;G(F^{-1}(z)),\sigma_2)\end{aligned}\end{equation}
whose $\sigma_1,\sigma_2$ are training parameters (both are scaler), $E(\cdot),G(\cdot)$ are training encoder and decoder, and $F(\cdot)$ is unconditional flow. Plug $\eqref{eq:f-vae-fxu}$ into $\eqref{eq:f-vae-loss}$, we get the final loss
\begin{equation}\begin{aligned}\iint \tilde{p}(x)q(u)\bigg[ &\frac{1}{2\sigma_2^2}\Vert G(\sigma_1 u + E(x))-x\Vert^2 \\
& + \frac{1}{2}\Vert F(\sigma_1 u + E(x))\Vert^2 - \frac{1}{2}\Vert u\Vert^2 \\
& - \log \left|\det \left[\frac{\partial F(\sigma_1 u + E(x))}{\partial u}\right]\right|\bigg] dudx\end{aligned}\end{equation}
The sampling procedure is
\begin{equation}u \sim q(u), \quad z = F^{-1}(u),\quad x = G(z)\end{equation}

\section{Related Work}

In fact, flow-based models are the general terms of a series of models. Except the above Coupling layer, we also have \emph{autoregressive flows}, whose representative works are PixelRNNs, PixelCNNs, and so on\cite{Oord2016Pixel,Salimans2017PixelCNN}. Autoregressive flows also work well, but they generate images pixel by pixel. Therefore autoregressive flows based models usually have a very slow speed.

Normalizing flows like RealNVP and Glow, are an another imporving choice of flows, especially Glow has shown amazing results in generation effect of flow-based models. Glow has a lower inferring time cost but training cost is still very heavy.

The first try to combine VAEs and flow-based models is \cite{Rezende2015Variational}. And a developed job is \cite{Chen2016Variational} and \cite{Kingma2017Improving}. They introduces (Inverse) Autoregressive Flows into VAEs. Both of these results (included ours) are similar. However, all the former jobs do not give a general framework like $\eqref{eq:f-vae-loss}$. And none of them achieve a breakthrough improvement on image generation.

Our work seems to be the first one to introduce (normalizing) flows like RealNVP and Glow into VAE. These flows are based on Coupling layer $\eqref{eq:coupling}$ and can be computed parallelly. So they are more efficient than autoregressive flows and can be stacked deeply. We also ensure that the latent variables dimensional are not compressed, thus alleviating the problem of generated blurry.

\section{Experiments}

\subsection{Procedure}

Limitted by GPUs, we can only evalute our model on 64x64 and 128x128 resolution of CelebA HQ. We firstly make a fast comparision with ordinary VAEs and Glow on 64x64 and then demonstrates high-quality generation on 128x128.

The encoder $E(\cdot)$ is stacks of Convolution and Squeeze operations. In details, our $E(\cdot)$ contains serveral blocks and apply Squeeze operation before echo blocks. And echo blocks contains serveral steps, denoted by $x + CNN(x)$, while $CNN(x)$ is stack of a 3x3 Relu Convolution and 1x1 linear Convolution. 

The decoder $G(\cdot)$ is stacks of Convolution and UnSqueeze, whose structure is the inverse of $E(\cdot)$. Add $\tanh (\cdot)$ activation at the final of $G(\cdot)$ is usually a good choice, but not necessary all the time. The structure of flow $F(\cdot)$ we use is also a multi-scale design like Glow, but with smaller depth and smaller kernel size of Convolution. 

\begin{figure*}
  \centering
  \setlength{\abovecaptionskip}{0.1cm}
  \subfigure[Ordinary VAEs]{
    \label{fig:vae}
    \includegraphics[height=3.8cm]{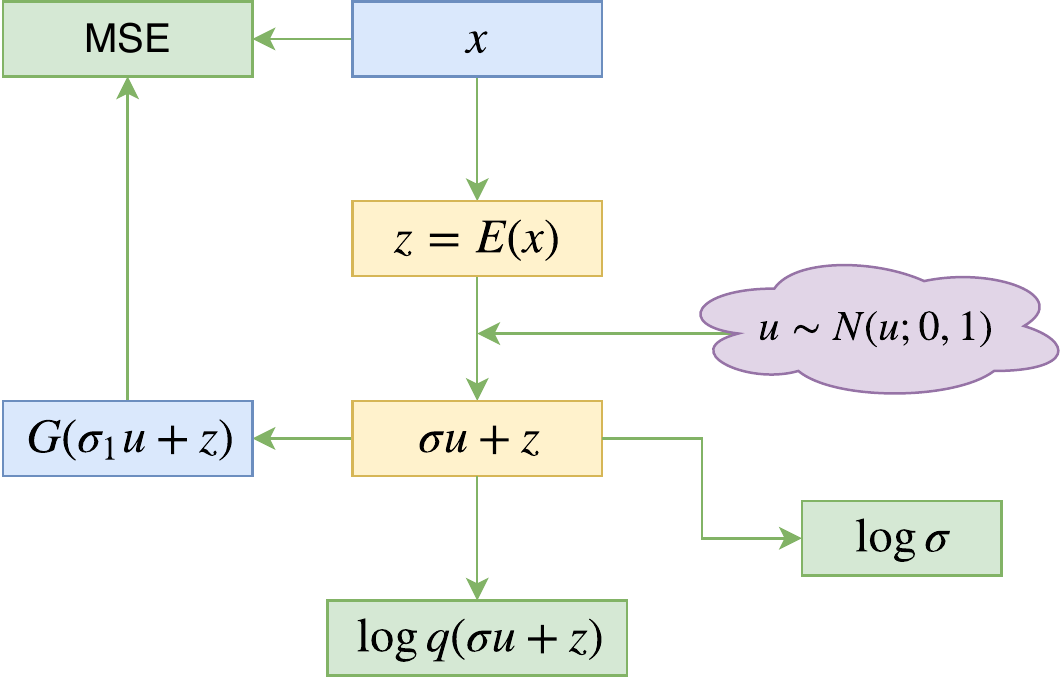}}\hspace{1cm}
  \subfigure[flow-based models]{
    \label{fig:flow}
    \includegraphics[height=3.8cm]{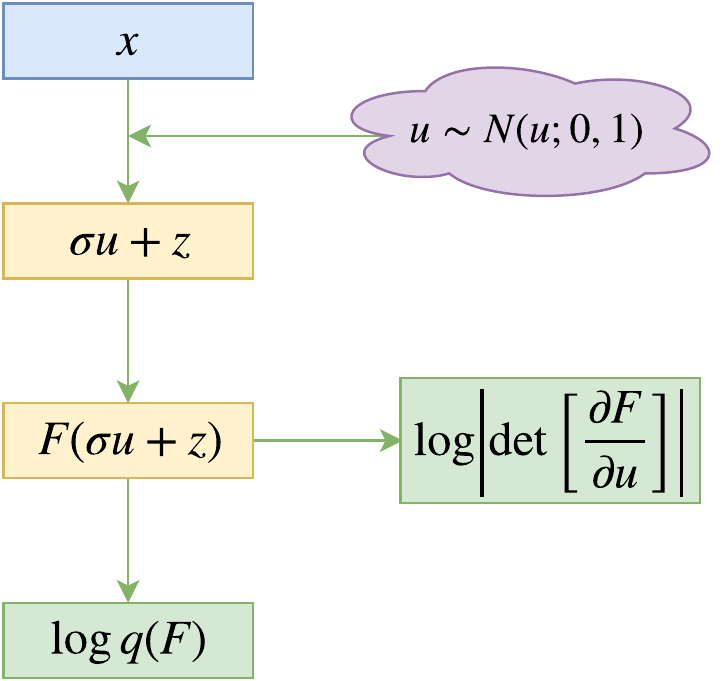}}\hspace{1cm}
  \subfigure[f-VAEs (ours)]{
    \label{fig:f-vae}
    \includegraphics[height=3.8cm]{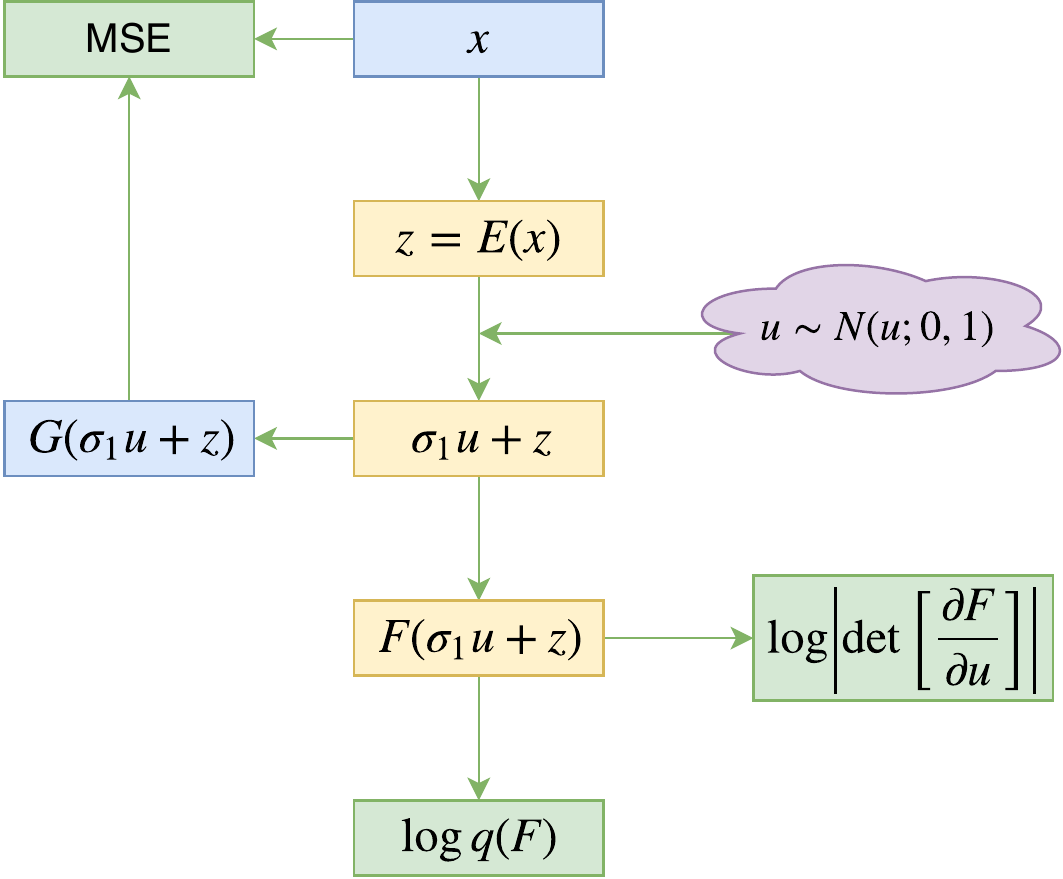}}
  \caption{Comparasion of Procedure of VAEs, flow and f-VAEs.}
  \label{fig:Comparasion}
\end{figure*}

\subsection{Results}

\begin{figure*}
  \centering
  \setlength{\abovecaptionskip}{0.1cm}
  \subfigure[Samples from VAEs]{
    \label{fig:samples-vae}
    \includegraphics[height=5.5cm]{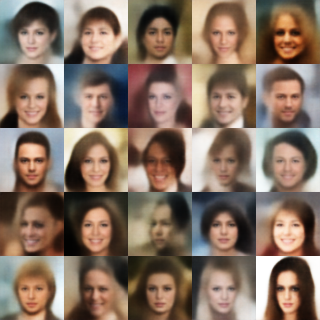}}\hspace{0.4cm}
  \subfigure[Samples from Glow]{
    \label{fig:samples-flow}
    \includegraphics[height=5.5cm]{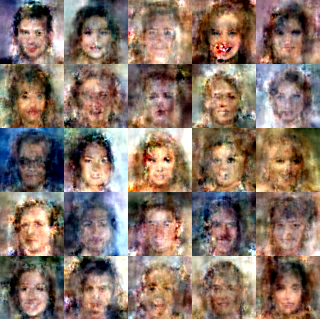}}\hspace{0.4cm}
  \subfigure[Samples from f-VAEs]{
    \label{fig:samples-f-vae}
    \includegraphics[height=5.5cm]{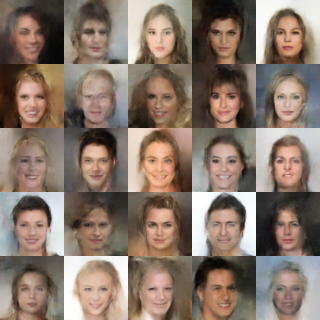}}
  \caption{Samples from VAEs, Glow and f-VAEs.}
  \label{fig:Comparasion2}
\end{figure*}

Compare figure $\ref{fig:samples-vae}$ and $\ref{fig:samples-f-vae}$, we can see f-VAEs has basically solved the blurring problem. We also compare the result of similar Glow on the same epoch (figure $\ref{fig:samples-flow}$). We do not doubt Glow will also perform well under more layers and more depths. But, obviously, f-VAEs perform better a lot than Glow under the same complexity and same training time. For achieving the above results, we only use one GTX 1060 to train about 7 hours (120 epochs).

Actually, we can see the total encoder of f-VAEs is $F(E(\cdot))$, which is composite of $F$ and $E$. In ordinary flow-based models, we need to calculate the Jacobi determinant of $E$, as it doest not need in f-VAEs. Therefore $E$ can be a general Convolutional Network, which can realize strong non-linearity.

\begin{figure*}
  \centering
  \setlength{\abovecaptionskip}{0.1cm}
  \includegraphics[height=5.5cm]{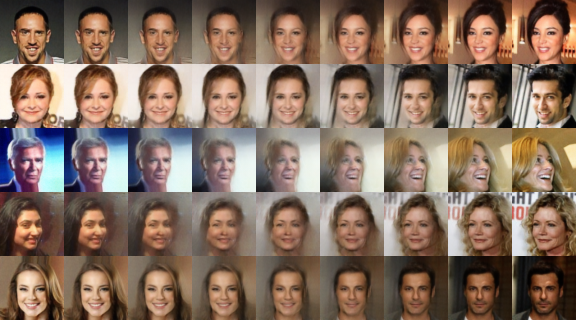}
  \caption{Linear interpolation in latent space of f-VAEs between real images}
  \label{fig:Interpolation}
\end{figure*}

Random interpolation in figure $\ref{fig:Interpolation}$ also show that encoder $F(E(\cdot))$ of f-VAEs transforms input images into a good embedding space.

Our results on 128x128 show in Appendix B.

\section{Discussion}

\subsection{Conclusion}

Actually, the orginal goal of this paper is to ask the following questions of Glow:

\begin{quote}\begin{enumerate}
\item How to reduce computation of Glow ?
\item How to implement a dimension-reduced version of Glow ?
\end{enumerate}\end{quote}

Our results show that a dimension-kept f-VAEs also equals a tiny-computation version of Glow but have better result. And $\eqref{eq:f-vae-loss}$ actually allow us to train a dimension-reduced flow-based model. We also reveal that ordinary VAEs and flow-based models are included in our framework theoretically. So we can say we get a more general generative and inference framework.

\subsection{Future Work}

we can see that random samples from f-VAEs still have a style like oil painting. One reason we think is our experimential model is not complex enough to fit the detail. But the most important reason we think is the abuse of 3x3 Convolution. It will make the perceptual field of Convolution infinitely expanded, leading the Convolution can not focus on the major details.

Therefore, a challenging task is to find out the magics of how to design workable and reasonable encoder and decoder. It seems that structures like Network in Network \cite{Lin2013Network} may be suitable to do that. But it is still waiting to be validated. And progressive growing structure like \cite{Karras2017Progressive} is also worth to have a try.

\newpage
\bibliographystyle{aaai}
\bibliography{biblio.bib}

\onecolumn
\appendix
\section{Appendix}

\subsection{A. Detailed Derivation of Equation $\eqref{eq:f-vae-loss}$\label{sec:derivation-loss}}

Combine $\eqref{eq:vae-loss}$ and $\eqref{eq:cond-flow}$, we have
\begin{equation}\label{eq:vae-loss-cond-flow}\begin{aligned}&\iiint \tilde{p}(x)\delta(z - F_x(u))q(u)\log \frac{\tilde{p}(x)\int\delta(z - F_x(u'))q(u')du'}{q(x|z)q(z)} dzdudx\\
=&\iint \tilde{p}(x)q(u)\log \frac{\tilde{p}(x)\int\delta(F_x(u) - F_x(u'))q(u')du'}{q(x| F_x(u))q(F_x(u))} dudx
\end{aligned}\end{equation}
Let $v = F_x(u'), u'=H_x(v)$, we have relation of Jacobi determinant
\begin{equation}\det \left[\frac{\partial u'}{\partial v}\right]=1\Big/\det \left[\frac{\partial v}{\partial u'}\right]=1\Big/\det \left[\frac{\partial F_x (u')}{\partial u'}\right]\end{equation}
And $\eqref{eq:vae-loss-cond-flow}$ becomes
\begin{equation}\begin{aligned}&\iint \tilde{p}(x)q(u)\log \frac{\tilde{p}(x)\int\delta(F_x(u) - v)q(H_x(v))\left|\det \left[\frac{\partial u'}{\partial v}\right]\right|dv}{q(x| F_x(u))q(F_x(u))} dudx\\
=&\iint \tilde{p}(x)q(u)\log \frac{\tilde{p}(x)\int\delta(F_x(u) - v)q(H_x(v))\Big/\left|\det \left[\frac{\partial F_x (u')}{\partial u'}\right]\right|dv}{q(x| F_x(u))q(F_x(u))} dudx\\
=&\iint \tilde{p}(x)q(u)\log \frac{\tilde{p}(x) q(H_x(F_x(u)))\Big/\left|\det \left[\frac{\partial F_x (u')}{\partial u'}\right]\right|_{v=F_x(u)}}{q(x| F_x(u))q(F_x(u))} dudx\\
=&\iint \tilde{p}(x)q(u)\log \frac{\tilde{p}(x) q(u)}{q(x| F_x(u))q(F_x(u))\left|\det \left[\frac{\partial F_x (u)}{\partial u}\right]\right|} dudx
\end{aligned}\end{equation}

\subsection{B. Results on 128x128}

\begin{figure*}
  \centering
  \setlength{\abovecaptionskip}{0.1cm}
  \includegraphics[height=7.5cm]{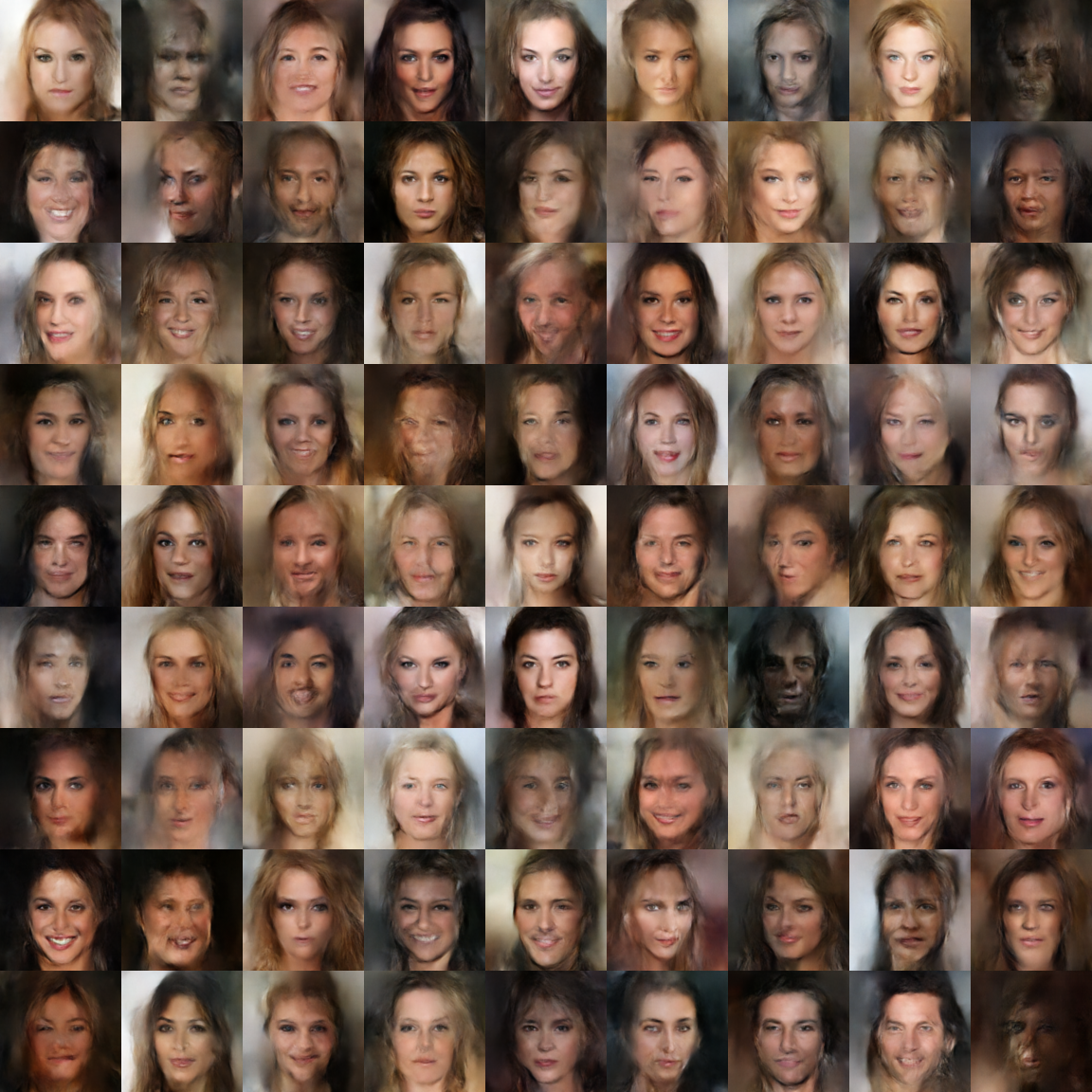}\hspace{0.4cm}
  \includegraphics[height=7.5cm]{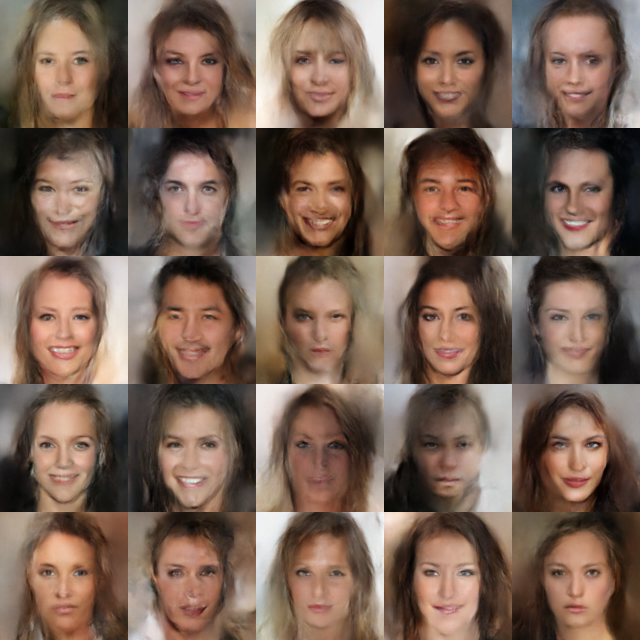}
  \caption{Samples from 128x128 model at temperature (standard deviation of prior distribution) 0.8}
  \label{fig:samples-128-T-0-8}
\end{figure*}

\begin{figure*}
  \centering
  \setlength{\abovecaptionskip}{0.1cm}
  \includegraphics[height=5.5cm]{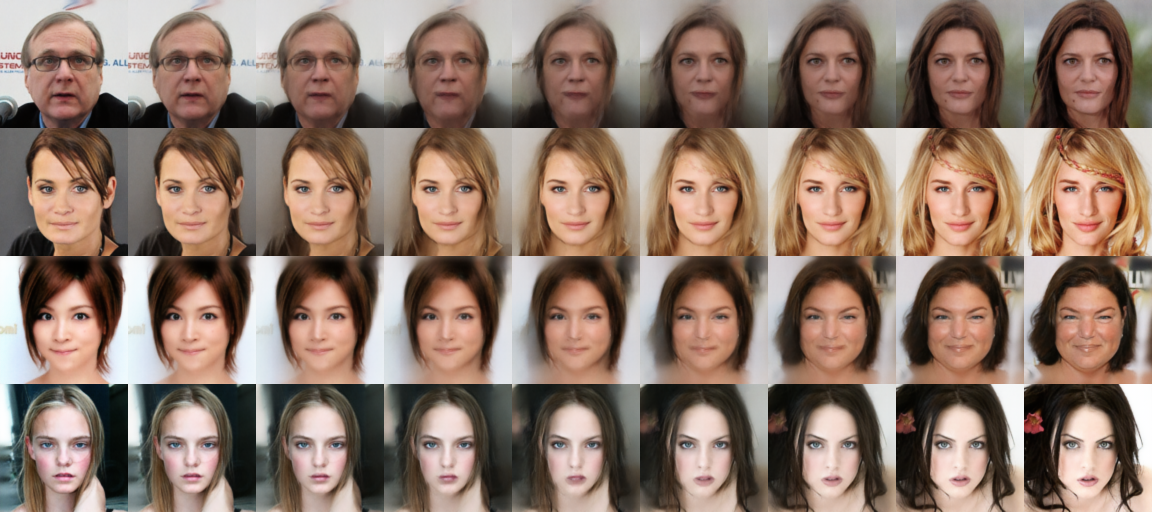}
  \caption{Linear interpolation in latent space between 2 real images}
  \label{fig:samples-128-interpolation}
\end{figure*}

\begin{figure*}
  \centering
  \setlength{\abovecaptionskip}{0.1cm}
  \includegraphics[height=6cm]{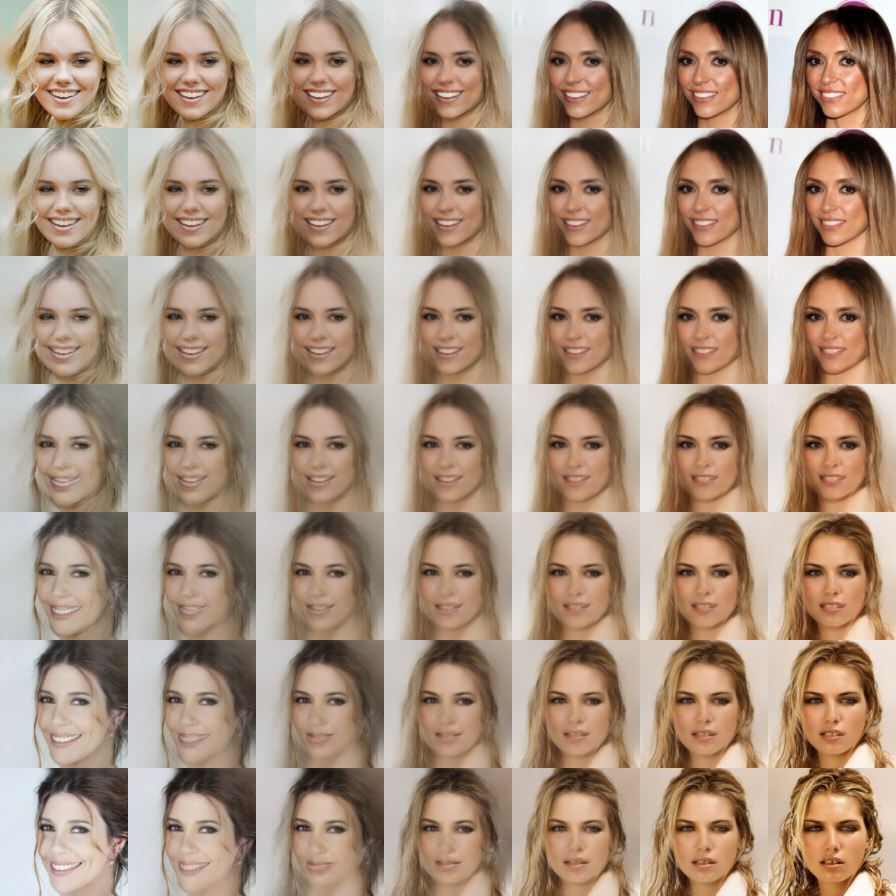}\hspace{0.4cm}
  \includegraphics[height=6cm]{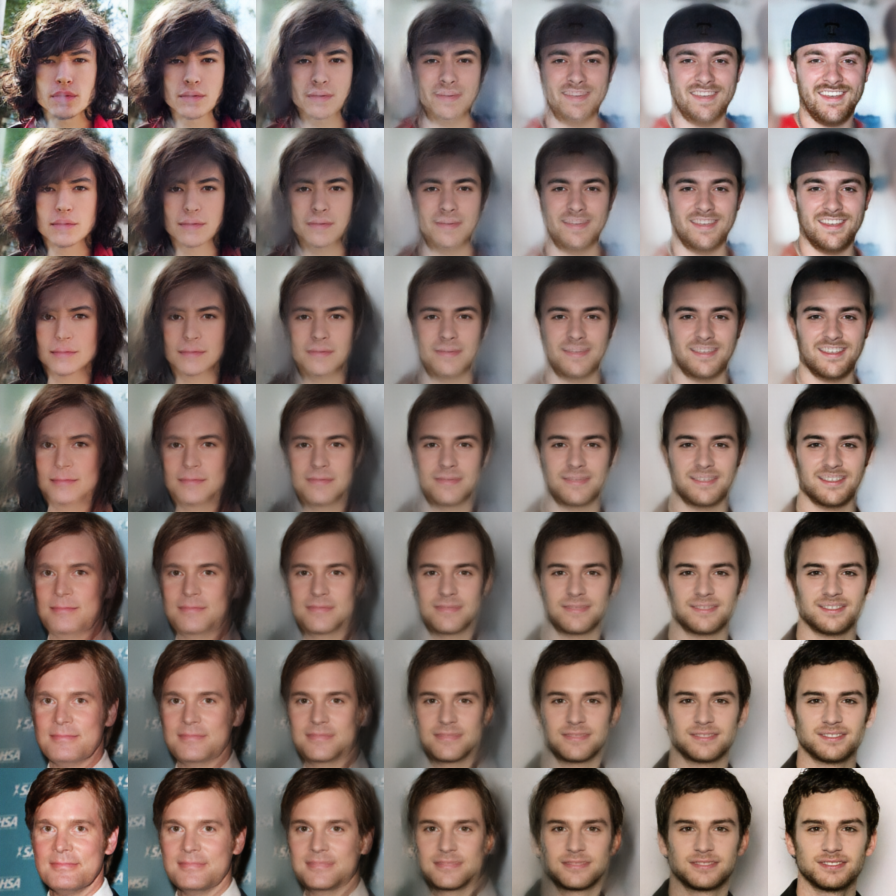}
  \caption{Linear interpolation in latent space between 4 real images}
  \label{fig:samples-128-interpolation-4}
\end{figure*}

\begin{figure*}
  \centering
  \setlength{\abovecaptionskip}{0.1cm}
  \subfigure[$T$=0]{
    \label{fig:samples-128-TT-0-0}
    \includegraphics[height=5cm]{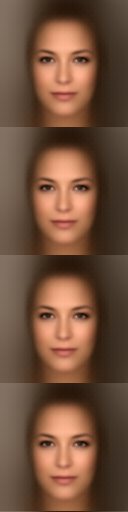}}\hspace{0.4cm}
  \subfigure[$T$=0.5]{
    \label{fig:samples-128-TT-0-5}
    \includegraphics[height=5cm]{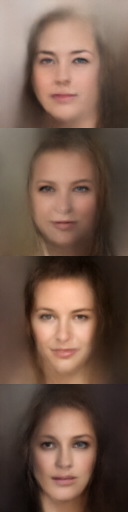}}\hspace{0.4cm}
  \subfigure[$T$=0.6]{
    \label{fig:samples-128-TT-0-6}
    \includegraphics[height=5cm]{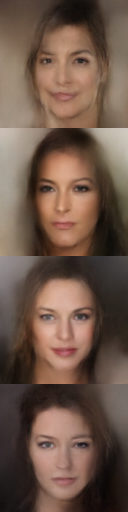}}\hspace{0.4cm}
  \subfigure[$T$=0.7]{
    \label{fig:samples-128-TT-0-7}
    \includegraphics[height=5cm]{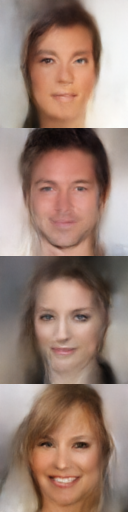}}\hspace{0.4cm}
  \subfigure[$T$=0.8]{
    \label{fig:samples-128-TT-0-8}
    \includegraphics[height=5cm]{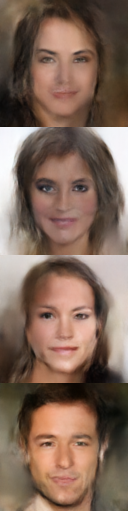}}\hspace{0.4cm}
  \subfigure[$T$=0.9]{
    \label{fig:samples-128-TT-0-9}
    \includegraphics[height=5cm]{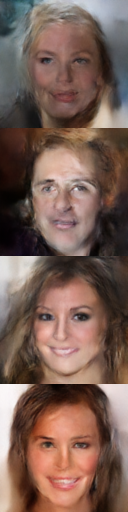}}\hspace{0.4cm}
  \subfigure[$T$=1.0]{
    \label{fig:samples-128-TT-1-0}
    \includegraphics[height=5cm]{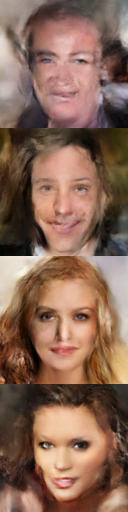}}
  \caption{Effect of change of temperature. From left to right, samples obtained at temperatures 0, 0.5, 0.6, 0.7, 0.8, 0.9, 1.0}
  \label{fig:samples-128-temperature}
\end{figure*}

We also validate our model on 128x128 CelebA HQ, whose results show in figure $\ref{fig:samples-128-T-0-8},\ref{fig:samples-128-interpolation},\ref{fig:samples-128-interpolation-4},\ref{fig:samples-128-temperature}$. We just trian it on one GTX1060, costing only about 1.5 day (150 epoch). 

\end{document}